\documentclass{article}
\usepackage{amsmath}
\usepackage{amssymb}
\usepackage{graphicx}
\usepackage{wrapfig}
\usepackage{xspace}
\usepackage{xcolor}
\usepackage{booktabs} 
\usepackage{bm}
\usepackage[final]{corl_2025} \usepackage{enumitem}

\definecolor{royalblue}{RGB}{65, 105, 225}
\definecolor{softgreen}{RGB}{85, 170, 85} \definecolor{softred}{RGB}{200, 50, 50}   
\hypersetup{
    colorlinks=true,
    linkcolor=orange,        citecolor=softgreen,        urlcolor=royalblue      }

\newcommand{\system}{RTR\xspace} 

\title{Robot Trains Robot: Automatic Real-World Policy Adaptation and Learning for Humanoids}

\author{
Kaizhe Hu$^*$ \quad
Haochen Shi$^*$ \quad
Yao He \quad
Weizhuo Wang \quad
C. Karen Liu$^\dagger$ \quad
Shuran Song$^\dagger$\\
$^*$Equal contribution \quad $^\dagger$Equal advising \\
 Stanford University \\
 \texttt{\href{robot-trains-robot.github.io}{robot-trains-robot.github.io}}
}

\begin{document}
\maketitle

\begin{abstract}
                    Simulation-based reinforcement learning (RL) has significantly advanced humanoid locomotion tasks, yet direct real-world RL from scratch or adapting from pretrained policies remains rare, limiting the full potential of humanoid robots. Real-world learning, despite being crucial for overcoming the sim-to-real gap, faces substantial challenges related to safety, reward design, and learning efficiency. To address these limitations, we propose Robot-Trains-Robot (\system), a novel framework where a robotic arm teacher actively supports and guides a humanoid robot student. The \system system provides protection, learning schedule, reward, perturbation, failure detection, and automatic resets. It enables efficient long-term real-world humanoid training with minimal human intervention. Furthermore, we propose a novel RL pipeline that facilitates and stabilizes sim-to-real transfer by optimizing a single dynamics-encoded latent variable in the real world. We validate our method through two challenging real-world humanoid tasks: fine-tuning a walking policy for precise speed tracking and learning a humanoid swing-up task from scratch, illustrating the promising capabilities of real-world humanoid learning realized by RTR-style systems. 
\end{abstract}

\keywords{Humanoid Robots, Sim-to-Real Adaptation, Real-World RL} 

\begin{wrapfigure}{r}{0.5\linewidth} 
    \vspace{-13mm}
    \centering
    \includegraphics[width=\linewidth]{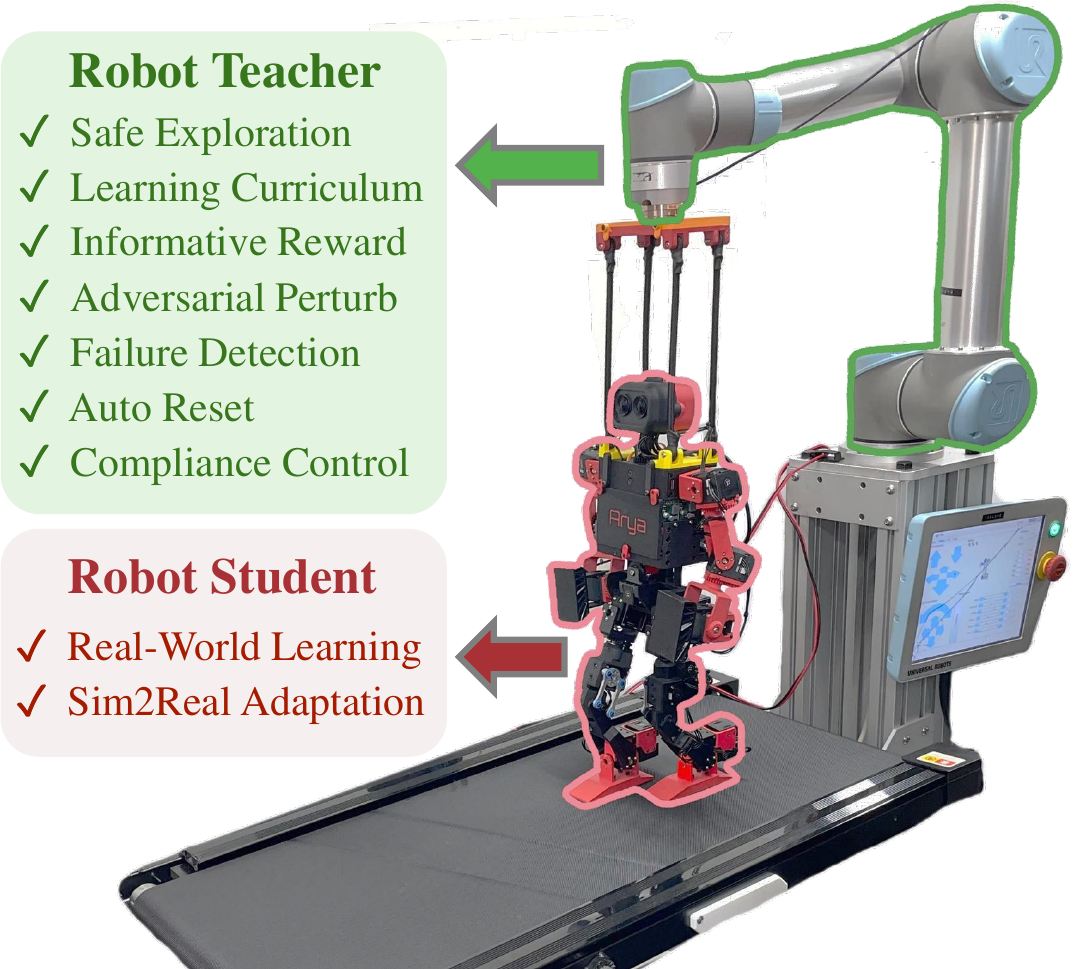}
    \caption{\textbf{Robot Trains Robot (RTR).} We propose \system for automatic real-world policy adaptation and learning with a robot arm as the teacher and a humanoid robot as the student.}
    \label{fig:teaser}
    \vspace{-5mm}
\end{wrapfigure}

\section{Introduction}

Recent advances in training reinforcement learning (RL) policies using massive parallel simulation environments have yielded remarkable results in humanoid locomotion tasks~\citep{he2025asap,he2025hover,radosavovic2024realworld,radosavovic2024learning,zhuang2024humanoid}. 
These methods demonstrate the ability to deploy on physical humanoid robots in a zero-shot manner or via real-world adaptation.

Nevertheless, learning directly in the real world remains arguably the most effective way to pursue optimal performance, as it bypasses the inevitable sim-to-real gap~\citep{tobin2017domain}. This is especially critical for complex systems like humanoid robots, where discrepancies between simulation and real-world dynamics can significantly hinder policy capability.
Despite these benefits, real-world learning on humanoids, either from scratch or fine-tuning, remains highly challenging due to several key factors:

\begin{itemize}[leftmargin=4mm]
\item \textbf{Safety}: Real-world RL can produce unexpected and dangerous actions during exploration, easily breaking the fragile balance of the humanoid. Current approaches typically suspend humanoid robots from fixed gantries during real-world rollout, inevitably constraining the range of data collection. Such passive protection mechanisms also influence learning efficiency, as the robot will be frequently dragged by the gantry and fall before the policy stabilizes.
\item \textbf{Reward Design}: Many useful reward signals easily available in simulation, such as global velocity or external forces, are difficult or impossible to measure directly on physical robots themselves, complicating real-world reward design. 

\item\textbf{Learning Efficiency}: Real-world training requires frequent resets, which incur high human labor costs. Poor sample efficiency of the learning algorithm further compounds these issues, as real-world data collection is inherently costly and constrained. 
\end{itemize}

These considerations highlight the need for a comprehensive and practical system tailored explicitly for real-world humanoid learning. To address these challenges, we propose a novel real-world policy adaptation and learning paradigm: \textbf{Robot-Trains-Robot (\system)}, where a teacher robot (a robot arm with force feedback) trains another student robot (a humanoid). Specifically:

\begin{itemize}[leftmargin=4mm]
    \item To address safety concerns, \system utilizes a robot arm with force-torque sensing to actively support and deliver interactive force feedback to the humanoid robot via compliance control, allowing extensive yet safe exploration across diverse behaviors.
    \item To gather crucial reward information, \system uses real-time measurements from the teacher to derive proxy reward signals that would otherwise be difficult to obtain in the real world.
    \item To improve learning efficiency, \system uses an automatic curriculum to provide procedural guidance, dynamically adjusting training difficulty, and deliberately perturbing the robot to enhance robustness. The \system system also enables failure detection, automatic resets, and an asynchronous data collection and policy update pipeline, significantly reducing manual intervention and enabling high-throughput data collection. 
\end{itemize}

Additionally, to further improve learning efficiency, we propose a novel \textbf{sim-to-real fine-tuning algorithm}. Our approach consists of a three-stage process: First, we train a dynamics-aware policy in simulation via domain randomization, embedding environment physics information into the policy via a latent encoder and FiLM~\citep{perez2017film} layers. Next, we optimize a universal latent vector across diverse simulated environments, providing a robust initialization for subsequent real-world training. Finally, leveraging the \system hardware, we efficiently refine the dynamics latent in the real world using PPO~\citep{schulman2017proximal}, substantially improving the policy performance and robustness in the real world. 

We evaluate our system and method through two challenging real-world humanoid tasks: fine-tuning a walking policy for precise speed tracking, and learning a humanoid swing-up behavior from scratch. Given a desired velocity command, \system effectively doubles the zero-shot walking speed with only $20$ minutes of real-world training. In the swing-up task, the humanoid successfully learns to achieve a periodical swing-up motion from scratch within $15$ minutes of real-world interaction. While we implement the \system system with an open-source small-sized humanoid ToddlerBot~\citep{shi2025toddlerbota}, the proposed paradigm is readily generalizable to full-scale humanoids. Contemporary industrial robotics arms are capable of lifting and interacting with payloads up to $ 600~\mathrm {kg}$~\citep{mir_mc600_2024}, suggesting the broad applicability of \system across a wide range of humanoid platforms.

In summary, our contributions include:
\textbf{(1)} A comprehensive real-world learning system that provides crucial protection, guidance, automation, and informative feedback tailored for real-world humanoid learning.
\textbf{(2)} An efficient RL paradigm leveraging dynamics-aware latent optimization for rapid and stable real-world policy adaptation.
\textbf{(3)} Empirical validation of our approach through two challenging tasks highlighting \system's efficiency and capability to support and generalize across diverse real-world learning scenarios.
\section{Related Works}

To address the sim-to-real gap~\citep{tobin2017domain} and improve policy performance in the real world, there are three mainstream approaches: zero-shot sim-to-real transfer, policy pretraining in simulation followed by real-world adaptation, and direct real-world RL from a random initialization.

\textbf{Zero-shot Sim-to-real Transfer.} Directly deploying a simulation-pretrained policy into the real world is prone to failures due to the sim-to-real gap.
Pretraining in simulation with extensive domain randomization has emerged as an effective strategy across both manipulation~\citep{peng2018simtoreal, tobin2017domain, yuan2024learning} and locomotion~\citep{ha2024umionlegs, rudin2022learning, cheng2024extreme, tan2018simtoreala, shi2023reference} tasks. This approach has notably succeeded even in challenging contexts such as dexterous manipulation involving rich contact interactions~\citep{openai2019solving, chen2023visual, chen2023sequential, chen2024objectcentric, lin2025simtoreal} and complex humanoid locomotion characterized by inherently unstable dynamics~\citep{fu2024humanplus, gu2024advancing, haarnoja2024learning, radosavovic2024realworld, zhuang2024humanoid}.
However, inappropriate domain randomization may produce policies that either fail to adapt effectively to real conditions or become overly conservative, thus limiting performance.

\textbf{Real-world Adaptation.} A popular approach to bridging the sim-to-real gap is to leverage real-world data. These methods typically rely on a robust pretrained policy to safely generate on-policy data, facilitating targeted exploration within task-specific distributions.
Some adaptation methods use real-world data to adjust simulation parameter distributions, aligning simulated policy behaviors more closely with real-world experiences~\citep{chebotar2019closing, ren2023adaptsim, huang2023what}. Other approaches employ online adaptation techniques to fine-tune latent representations~\citep{schoettler2020metareinforcement, kumar2021rma, qi2022inhand, kumar2022adapting}, a residual policy~\citep{sun2021online, he2025asap}, or the original policy directly ~\citep{zhang2023cherrypicking, lei2023unio4, xiong2024adaptive}. Additionally, some studies incorporate online human corrections for policy refinement~\citep{jiang2024transic, luo2025precise}.
Notably, due to humanoids' inherent instability, it is challenging to maintain safety during real-world adaptation. Consequently, fewer studies focus on humanoids~\citep{he2025asap, kumar2022adapting}, highlighting the need for a specialized system to support safe and effective humanoid adaptation. 
Our proposed dynamics latent tuning pipeline is the most similar to context-based meta-RL methods~\citep{gupta2018meta, rakelly2019efficient}, where a latent vector is extracted from past observations or physical parameters and used to pivot the policy to adapt to different tasks or environments during meta-pretraining. The latent could then be optimized in the real world for rapid policy adaptation. While these methods have shown success in quadruped locomotion tasks~\citep{yu2020learning, peng2020learning}, they implicitly rely on the inherent stability of quadrupeds to safely initiate real-world interaction. For humanoids, any initial performance gap risks falls or hardware damage. Although early attempts have deployed such methods on humanoids~\citep{yu2019sim}, they rely on a simple platform and a scripted policy for initial data collection, which is impractical for more complex humanoid systems. In contrast to previous works, \system combines a carefully designed hardware setup with a simulation-optimized initial latent, enabling the real-world application of this category of algorithms on humanoid robots.

\textbf{Real-world RL.} For tasks that are difficult to simulate or have a large sim-to-real gap, directly training in the real world is often more desirable. Some studies have successfully adopted this approach across various setups, including tabletop manipulation~\citep{huang2024mentor}, dexterous manipulation~\citep{xu2023dexterous}, mobile manipulation~\citep{mendonca2025continuously}, and quadruped locomotion~\citep{ha2021learning, smith2022walk, wu2023daydreamer, smith2023grow}.
However, few studies have explored direct real-world RL with humanoids due to several unresolved challenges: (1) maintaining humanoid stability and preventing falls during early stages of training, (2) ensuring safe and efficient reset mechanisms, and (3) providing sufficient feedback and informative reward signals. One notable work of this kind by \citet{bloesch2022towards} use a much simpler humanoid platform ~\cite{robotis_op3_manual} with 20 DoFs to move forward via end-to-end reinforcement learning. We propose \system to address these challenges and enable real-world reinforcement learning  on a much more complex humanoid platform.

\section{Method}

\subsection{Teacher-Student Hardware System}
\begin{figure}[t]
    \centering
    \includegraphics[width=0.9\linewidth]{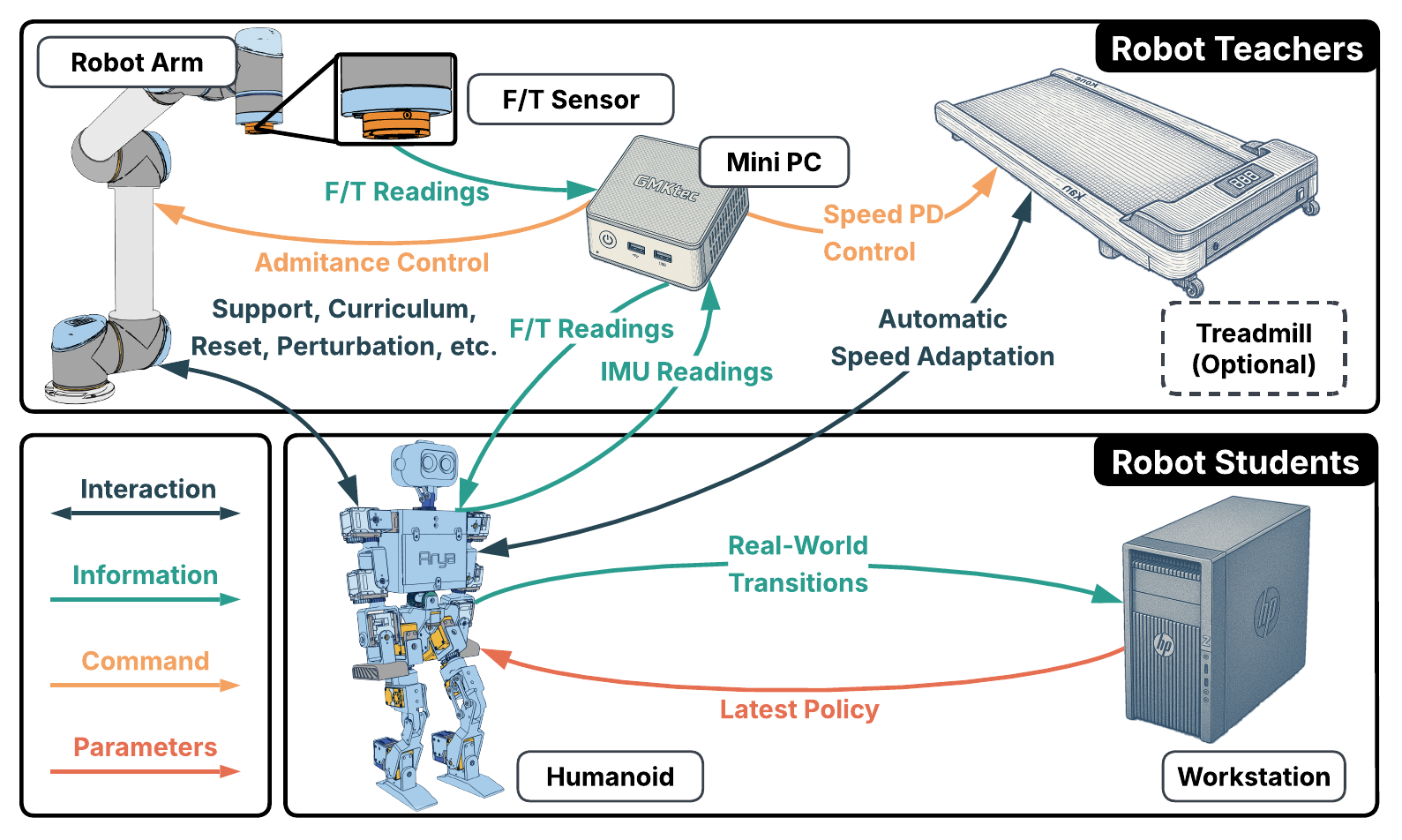}
    \caption{\textbf{System Setup.} We illustrate the system architecture and component interactions. The system consists of two groups: robot teachers and robot students. The teachers include a robot arm with an F/T sensor, a mini PC, and an optional treadmill for locomotion tasks; the students include a humanoid robot and a workstation for policy training. The four types of lines represent physical interaction, data transmission, control commands, and neural network parameters, respectively.}     \label{fig:system_setup}
    \vspace{-3mm}
\end{figure}

\textbf{Teacher Setup.} We use a 6-DoF UR5 arm to support the humanoid robot. An ATI mini45 force-torque (F/T) sensor is mounted on the arm's end-effector to measure interaction forces. Four elastic ropes connect the arm's end effector to the humanoid's shoulders. The elasticity of the rope is crucial, as it enables smoother force transmission and avoids abrupt force changes commonly seen in rigid connections or non-elastic ropes. For walking experiments, we additionally provide a programmable treadmill to ensure the robot stays within the reach of the robot arm. This treadmill is equipped with a position encoder and a microcontroller to provide closed-loop control of the moving speed. As shown in Figure~\ref{fig:system_setup}, a mini PC connects to the arm, F/T sensor, and treadmill via Ethernet cables, while communicating with the student robot via WiFi. This mini PC bears several functionalities: (1) sending control signals to the arm and the treadmill to ensure they assist the learning of the humanoid properly, and (2) collecting data from the F/T sensor and the treadmill to tailor the curriculum, reset timing, and gather reward information for the robot.

\textbf{Student Setup.} We use the open-source humanoid ToddlerBot~\citep{shi2025toddlerbota} for its compact size ($0.56~\mathrm{m}$, $3.4~\mathrm{kg}$), dexterity (30 degrees of freedom), availability (costs less than $6,000$ USD), and robustness. The UR5 arm, with a $5~\mathrm{kg}$ payload, provides adequate capacity to support the robot. As safety is a major concern in real-world learning, verifying our algorithm on a lightweight yet versatile platform like ToddlerBot enables unattended operation without risk of damaging itself or the surrounding environment. Moreover, hardware reliability for the humanoid is equally important—ToddlerBot's motors are sufficiently resistant to overheating and capable of continuous operation over extended periods ($>1$ hour), making the robot well-suited for our real-world learning tasks.

\subsection{Real-world Adaptation}

\begin{figure}[t]
    \centering
    \includegraphics[width=0.95\linewidth]{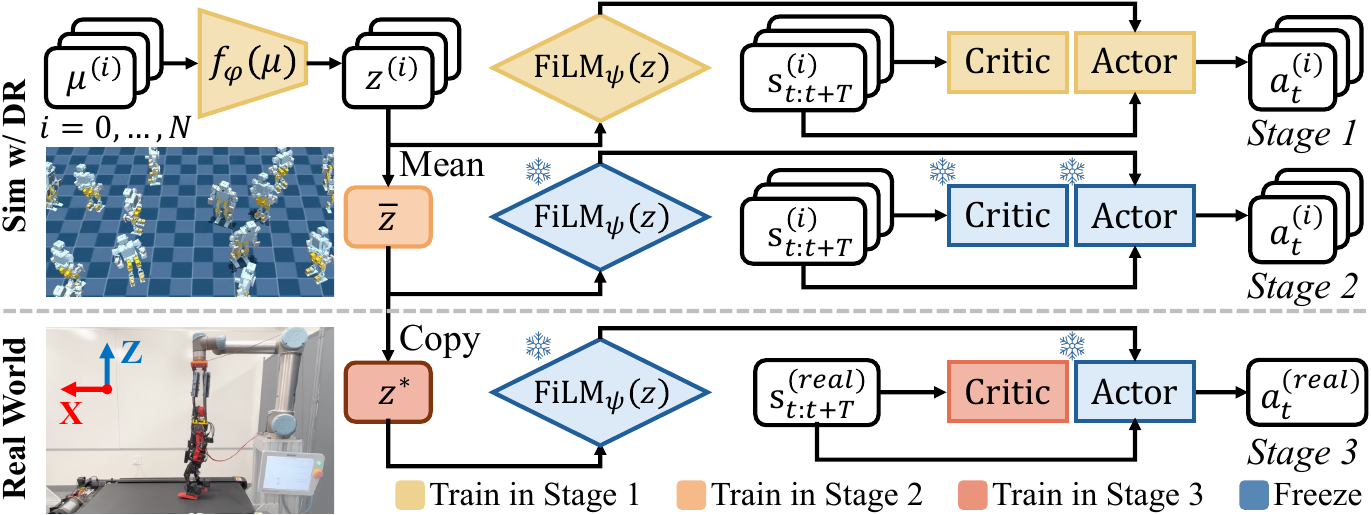}
    \caption{\textbf{Sim-to-real Fine-tuning Algorithm.} We illustrate our sim-to-real finetuning process. First, we train a dynamics-aware policy in simulation via domain randomization (DR), encoding environment physics into a latent vector. Next, we optimize a universal latent across diverse simulation environments to initialize real-world training. Finally, we refine the latent and train a new critic in the real world. Orange denotes trainable components in three stages; blue indicates frozen ones.} 
            \label{fig:adaptation_method}
    \vspace{-3mm}
\end{figure}

Real-world adaptation, which fine-tunes a simulation-pretrained policy in the real world, is critical for improving performance on demanding locomotion tasks. 
In this section, we apply our real-world adaptation pipeline to fine-tune a walking policy for precise speed tracking, while our setup is suitable for adapting a range of locomotion tasks like running and whole-body trajectory following. 

On the teachers' side, we introduce an automatic curriculum that dynamically adjusts the supporting force and enables rapid horizontal compliant arm following. On the students' side, we propose a general-purpose sim-to-real adaptation algorithm based on domain randomization and dynamics latent optimization.
We first present our three-stage student learning algorithm, as illustrated in Figure~\ref{fig:adaptation_method}, and then describe the teacher’s policy during real-world adaptation.

\textbf{Dynamics-conditioned Policy Training.} 
In the first stage, we train a dynamics-aware policy $\pi(s, z)$ that conditions on both the current observation $s$ and a dynamics latent $z$ in $N=1000$ domain randomized simulation environments (more details in Appendix~\ref{app:repr}). Specifically, for the $i$-th environment, we encode environment-specific physical parameters $\mu^{(i)}$ into a latent representation $z^{(i)}$ using a multi-layer perceptron (MLP) encoder $f_\phi$: $z^{(i)} = f_\phi(\mu^{(i)})$. This dynamics latent is then incorporated into the standard PPO actor network through Feature-wise Linear Modulation (FiLM) layers \citep{perez2017film}. Concretely, let $h_j^{(i)}$ denote the hidden latents of the $j$-th layer within the actor network after activation function, we modulate these latents using FiLM as follows:

\begin{equation}
\label{eq:film_layer}
\gamma_j^{(i)}, \beta_j^{(i)} = \text{FiLM}_j(z^{(i)}),\quad
h_j^{(i)} \leftarrow \gamma_j^{(i)} \odot h_j^{(i)} + \beta_j^{(i)},
\end{equation}

where $\gamma_j^{(i)}$ and $\beta_j^{(i)}$ are scaling and shifting parameters generated by the FiLM layers from the latent vector $z^{(i)}$, and $\odot$ represents element-wise multiplication. The encoder network $f_\phi$ and the FiLM-modulated policy network $\pi(s,z)$ are jointly trained with PPO in randomized domains with parameters $\mu^{(i)}$, allowing the policy to leverage latent dynamics information effectively and adapt to various simulation environments. Empirically, we find that the learning rate of the FiLM layers is critical: a rate that is too small causes the policy to ignore the dynamics latent, while a rate that is too large leads to instability. An ablation study of this effect is provided in Appendix~\ref{app:film}. 
\textbf{Universal Latent Optimization.} A practical challenge arises when deploying the dynamics-aware policy in the real world: the initial latent representation $z$ for the real world is unknown, as the environment-specific parameters $\mu$ are unavailable. Therefore, we propose the second stage to optimize a universal latent $\tilde{z}$ starting from $\bar{z}$, the average of all the latent vectors of the training environments. Formally, we freeze the policy network and FiLM layer parameters and optimize $\tilde{z}$ using PPO, aiming for robust performance across all domain-randomized simulation environments:

\begin{equation}
\tilde{z} = \arg\max_z \sum_i \mathbb{E}_{\tau \sim \pi(\cdot \mid z),\, \mathcal{T}_i} [J(\tau)]
\end{equation}

where $\mathcal{T}_i$ denotes the transition distribution for the $i$-th environment, $J$ denotes the optimization objective of the PPO algorithm, i.e., the expected cumulative reward under policy $\pi$, and the expectation is taken over the joint distribution induced by the policy and the environment dynamics. The resulting latent $\tilde{z}$ provides a robust initial condition suitable for real-world deployment. 

\textbf{Real-world Finetuning.} In this stage, we freeze the actor network and FiLM layer parameters and fine-tune the latent in the real world using $\tilde{z}$ as the initial solution. Meanwhile, we train the critic from scratch since some privileged observations are unavailable in the real world, resulting in a different observation space from the simulation. For the walking task, the reward is designed to encourage tracking of the target velocity and is defined as follows:
\begin{equation}
r = \exp\left( -\sigma \cdot (v - v^{\text{target}})^2\right),
\label{eq:walk_reward}
\end{equation}
where $v$ and $v^{\text{target}}$ are the current and target robot velocity, respectively, and $\sigma=100$ is a reward-shaping hyperparameter. Since the humanoid is walking on a treadmill, its torso remains relatively stationary in the global frame - state estimation or motion capture yields a near-zero velocity. Therefore, we approximate $v$ with the speed of the treadmill.

\textbf{Real-world Teachers.} The robot teachers include the robot arm with an F/T sensor, a mini PC, and a treadmill. They together provide guidance, schedule, reward, failure detection, and automatic resets. \textbf{(1) Guidance:} Leveraging feedback from the F/T sensor, we employ admittance control~\citep{maples1986experiments, hou2025adaptive} for the robot arm. The arm remains compliant along the XY axes to accommodate the relative movement generated by the humanoid (Figure~\ref{fig:adaptation_method}). This configuration allows the humanoid to move freely in the XY direction while maintaining an upright posture.  \textbf{(2) Schedule:} To gradually reduce assistance, we implement a scheduling strategy where the arm's height linearly decreases by $0.02$m in $5\times10^4$ environment steps, diminishing the supporting force to near zero at the end of training.
\textbf{(3) Reward:} We also implement a PD feedback control loop on the treadmill's velocity, based on force along the X axis by the F/T sensor and the humanoid's torso pitch angle. This feedback loop helps the robot maintain an upright walking posture and ensures that the treadmill speed reflects the humanoid's walking speed. We further use this tracking speed to provide reward signal as in Equation~\eqref{eq:walk_reward}. \textbf{(4) Failure Detection and Automatic Resets:} Moreover, \system automatically detects failure if the humanoid’s torso pitch exceeds a threshold or the F/T sensor reads a large force along the X or Y axis,  prompting the system to step and the arm to lift the humanoid to reset the training.

\subsection{Real-world Learning from Scratch}
\label{sec:real_world_learning}
While sim-to-real learning is effective for humanoid locomotion, it struggles with tasks involving hard-to-simulate objects like deformable ones. \system is also suitable for direct real-world training in such cases. To demonstrate this flexibility, we introduce a challenging real-world RL task—learning a swing-up behavior (Figure~\ref{fig:swing_ablation})—which is difficult to simulate due to complex cable dynamics.

\textbf{Three-stage Training.}  Inspired by~\citet{lei2023unio4}, we use a three-stage training pipeline: (1) train the actor and critic from scratch in the real-world using PPO, and collect $50{,}000$ steps of suboptimal transition data; (2) pre-train the critic using offline RL on these data; and (3) initialize a new actor while loading the pretrained critic, and continue to train both networks jointly. The reward is designed to maximize the amplitude of the dominant periodic force measured by the force sensor during the swinging. Specifically, we applied the Fast Fourier Transform (FFT) on the most recent $1,000$ force readings along the x-axis to obtain the force spectrum in the frequency domain and retrieve the dominant frequency, denoted as $\nu_x$. We then extract the force amplitude at this frequency, $\hat{A}_{\nu_x}$, and define the reward as:
\begin{equation}
r = \exp\left(-\alpha \cdot (\hat{A}_{\nu_x} - A^{\text{target}})^2\right),
\end{equation}
where $\alpha=0.005$ and $A^{\text{target}} \approx m g \theta_0$ is derived under the small-angle approximation of a pendulum swing. $m = 3.5~\mathrm{kg} $ is the mass and $ \theta_0 = 30^\circ $ is the maximum angular displacement expected.

\textbf{Real-world Teachers.} The robot teacher guides the humanoid's swing-up motion by either amplifying the swing or damping the swing. Both strategies are phase-aligned with the humanoid's current swing angle $\theta_t$ at the dominant force frequency $\nu_x$. \textbf{(1) Guidance:} To amplify the swing, i.e., increase the force amplitude at the dominant frequency, the arm is given a position target $x_t = x_0 + A_{\text{arm}}\cos(\theta_t)$. \textbf{(2) Perturbation:} To dampen the swing, i.e., decrease the force amplitude, the arm is given a phase-inverted position target $x_t = x_0 - A_{\text{arm}}\cos(\theta_t)$, where $A_{\text{arm}} = 0.05~\mathrm{m}$, and $x_0$ denotes the initial $x$-axis displacement of the arm. \textbf{(3) Schedule:} We evenly divide each data batch into several bins and randomly select a certain number of bins to apply either guidance or perturbation. The arm maintains a fixed position in the rest bins. The final training schedule includes a mixture of helping, perturbing, and remaining static periods. 

\section{Experiments}
\label{sec:result}

\subsection{Real-world Adaptation with Simulation Pretraining: Walk}
\begin{figure}[t]
    \centering
    \includegraphics[width=\linewidth]{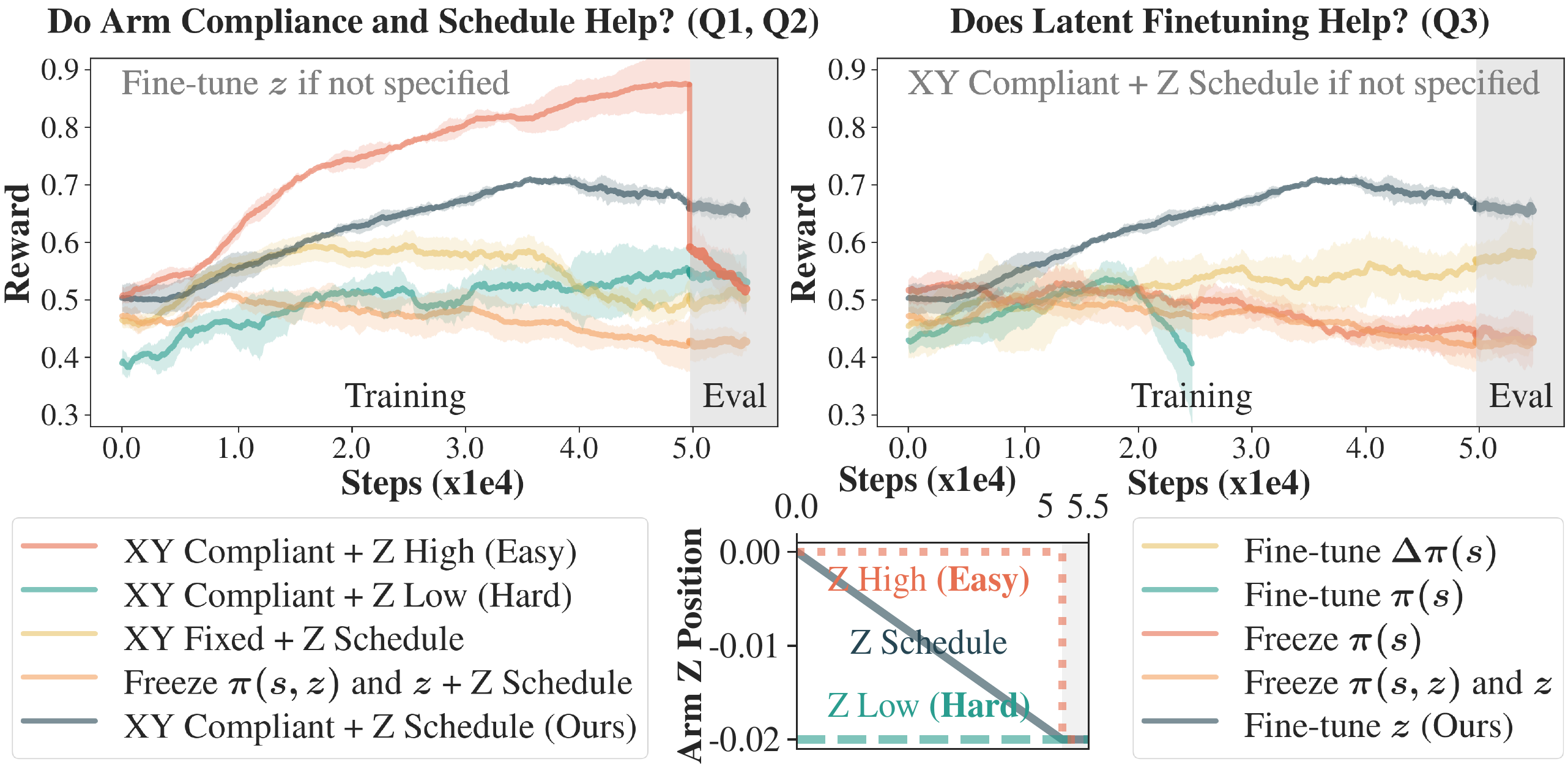}
    \caption{\textbf{Walking Ablation.} This experiment aims to evaluate the effectiveness of arm feedback control and latent vector finetuning. We present the linear velocity tracking rewards during training and evaluation, with the arm schedule shown at the bottom center. All variants are tested under the same condition: the arm uses an XY Compliant policy with Z fixed at a $\Delta$ position of $-0.02~\mathrm{m}$. We conduct each experiment with three random seeds and show the mean and standard deviation in the plot. Unless otherwise specified, \textit{Finetune} $z$, \textit{XY Compliant}, and \textit{Z Schedule} are assumed.}
    \label{fig:walk_ablation}
    \vspace{-2mm}
\end{figure}
\begin{table}[t]
    \centering
    \caption{We evaluate the walking baselines in this table. The humanoid's task is to track the treadmill speed ($0.15~\mathrm{m/s}$). To assess walking stability, torso pitch and roll (radian) are measured by the humanoid’s IMU, and end-effector (EE) force (N) by the F/T sensor. Unless otherwise specified, \textit{Finetune} $z$, \textit{XY Compliant}, and \textit{Z Schedule} are assumed if not otherwise specified.}
    \resizebox{\linewidth}{!}{
    \begin{tabular}{lccccc}
\toprule
Method &     Torso Roll $\downarrow$ &     Torso Pitch $\downarrow$ &     EE Force X $\downarrow$ &     EE Force Y $\downarrow$ &     EE Force Z $\downarrow$ \\ 
\midrule
Freeze $\bm{\pi(s)}$ & 
  0.124 $\pm$   0.012 &   0.102 $\pm$   0.034 &   1.217 $\pm$   0.134 &   1.235 $\pm$   0.065 &   3.431 $\pm$   0.606 \\
Finetune $\bm{\Delta\pi(s)}$ & 
  0.102 $\pm$   0.032 &   0.096 $\pm$   0.031 &   1.007 $\pm$   0.025 &   1.078 $\pm$   0.136 &   2.316 $\pm$   0.486 \\
Freeze $\bm{\pi(s, z)}$ & 
  0.110 $\pm$   0.021 &   0.064 $\pm$   0.014 &   1.672 $\pm$   0.174 &   1.088 $\pm$   0.186 &   3.654 $\pm$   0.988 \\
Z Fixed (High) & 
  0.151 $\pm$   0.030 &   0.126 $\pm$   0.011 &   1.064 $\pm$   0.143 &   0.993 $\pm$   0.279 &   2.402 $\pm$   0.732 \\
Z Fixed (Low) & 
  0.199 $\pm$   0.113 &   0.112 $\pm$   0.053 &   1.254 $\pm$   0.133 &   0.882 $\pm$   0.104 &   3.237 $\pm$   1.196 \\
XY Fixed & 
  0.170 $\pm$   0.030 &   0.156 $\pm$   0.027 &   1.175 $\pm$   0.130 &   0.864 $\pm$   0.100 &   2.299 $\pm$   0.820 \\

\system(ours) & 
\textbf{0.093 $\pm$   0.020} & \textbf{0.053 $\pm$   0.044} & \textbf{0.943 $\pm$   0.202} & \textbf{0.754 $\pm$   0.122} & \textbf{0.954 $\pm$   0.445} \\
\bottomrule
    \end{tabular}
    }
    \label{tab:walk_metrics}
    \vspace{-3mm}
\end{table}

\textbf{Overview.} We consider the task of walking on a treadmill while accurately tracking the treadmill’s speed to demonstrate \system’s real-world adaptation capability. During training and evaluation, the reward is defined as the treadmill’s speed, with the treadmill speed controlled via feedback from IMU and force readings. During testing, performance is measured by walking stability at a fixed treadmill speed of $0.15~\mathrm{m/s}$ (Table~\ref{tab:walk_metrics}). We conduct ablation studies to answer three key questions: \textbf{(Q1)} Does arm compliance control help? \textbf{(Q2)} Does arm schedule help? \textbf{(Q3)} Is fine-tuning $z^*$  more data efficient in real-world adaptation?

\textbf{Arm Compliance.} As shown on the left of Figure~\ref{fig:walk_ablation}, we compare \system with the baseline \textit{XY Fixed + Z Schedule}, where the arm remains stationary and cannot adapt to the humanoid’s XY movement during walking. We observe that the arm often drags the humanoid back and impairs policy adaptation. During evaluation, compliance control is re-enabled to ensure a fair comparison.

\textbf{Arm Schedule.} We compare our linearly decreasing arm height schedule with two alternative scheduling strategies: \textit{XY Compliant + Z High (Easy)} and \textit{XY Compliant + Z Low (Hard)}. The different arm schedules are illustrated in the lower center of Figure~\ref{fig:walk_ablation}. \textit{Z High (Easy)} trains the humanoid at a relatively high delta arm height ($0~\mathrm{m}$) and evaluates it at a low delta arm height ($-0.02~\mathrm{m}$). As shown on the left of Figure~\ref{fig:walk_ablation}, the training curve goes up very quickly during training because the robot takes advantage of the arm support, and then it collapses during evaluation when the arm height is lower due to the large training and evaluation gap. For \textit{Z Low (Hard)}, the training and evaluation arm heights are the same, so the policy should ideally overfit to the arm height. However, due to the high initial task difficulty, the policy frequently falls in the early stage, leading to poor data quality for finetuning and slower reward improvement compared to our method.

\textbf{Fine-tuning Latent.} We compare \system with two fine-tuning strategies: fine-tuning $\pi(s)$ and fine-tuning $\Delta \pi(s)$ (Figure~\ref{fig:walk_ablation}). Directly fine-tuning $\pi(s)$ initially progresses well but quickly collapses. In fine-tuning $\Delta \pi(s)$, the base policy $\pi(s)$ is frozen, and a residual policy is initialized and trained from $\pi(s)$ with the last layer weights reset to zero. We find that \system achieves better data efficiency than both baselines. All methods use the same arm policy, \textit{XY Compliant + Z Schedule}. For reference, we also show that without fine-tuning $\pi(s)$ or $\pi(s,z)$, the reward gradually decreases as the arm lowers and the task difficulty increases.
We additionally compare RTR with RMA~\citep{kumar2021rma} in Appendix~\ref{app:rma}.

\subsection{Real-world Learning from Scratch: Swing-up}

\textbf{Overview.} To demonstrate the effectiveness of \system for real-world learning from scratch, we consider training a swing-up behavior with RL in the real world. The objective is to achieve the maximum swing height within a 20-second time window. To simplify the task, we constrain the action space to hip pitch, knee pitch, and ankle pitch and enforce symmetry between the left and right legs. We perform ablation experiments to answer two key questions: \textbf{(Q1)} Does active arm involvement help in the training schedule? \textbf{(Q2)} Does critic pretraining help? 

\begin{figure}[t]
    \centering
    \includegraphics[width=\linewidth]{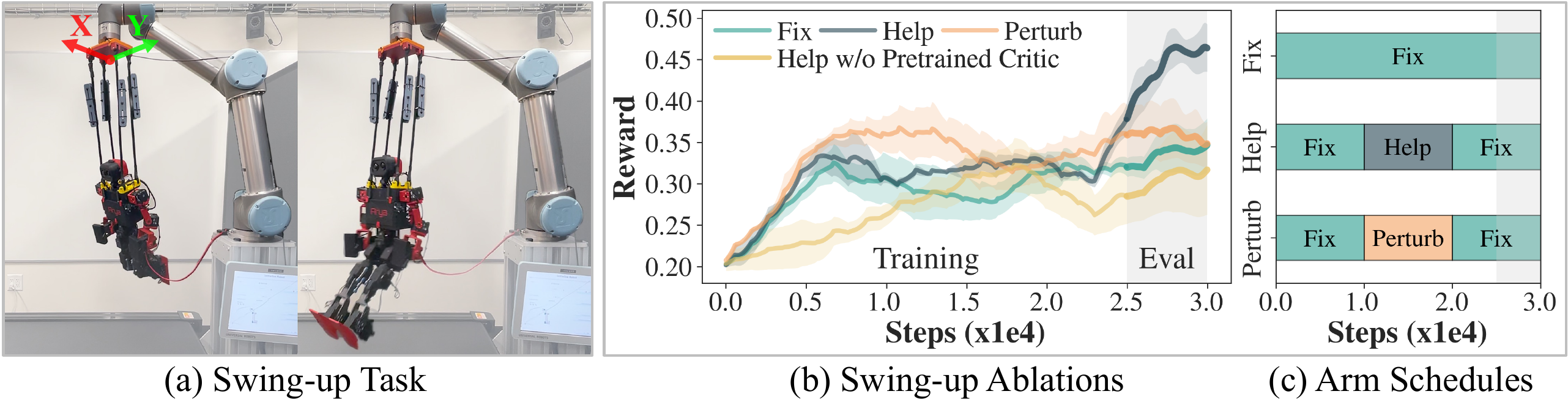}
    \caption{\textbf{Swing-up Ablation.} We illustrate the swing-up setup and experiment results. (a) The humanoid is suspended from a robot arm and uses its legs to build momentum and maximize rope angle. (b) We compare helping and perturbing arm schedules against a fixed-arm baseline and also evaluate helping without a pretrained critic. Each experiment is run with three random seeds, and plots show the mean and standard deviation. (c) We show three arm schedules, where helping and perturbing occur during the middle phase, with the arm fixed at the beginning and end.}
    \label{fig:swing_ablation}
    \vspace{-3mm}
\end{figure}

\textbf{Analysis.} As shown in Figure~\ref{fig:swing_ablation}, both the helping and perturbing arms outperform the fixed arm, with the helping achieving the best performance. We conclude that helping allows the critic to learn what good states look like during the helping interval ($1\text{e}4$ – $2\text{e}4$ steps), enabling the policy to quickly reach those states afterward. Pretraining the value function with offline data also accelerates early-stage learning - the model without a pretrained critic exhibits the slowest improvement.
\section{Conclusion}
\label{sec:conclusion}

We present a comprehensive real-world learning system, \system, which enables protection, guidance, schedule, reward,  perturbation, failure detection, and automatic resets for humanoid learning. We introduce an efficient RL paradigm based on dynamics-aware latent optimization, enabling rapid and stable policy adaptation in the real world. Looking ahead, we aim to extend \system to larger humanoid robots, handle increasingly complex tasks, and further improve data efficiency, allowing more effective collection and utilization of real-world data for humanoid learning.
\newpage

\section{Limitation}
Although RTR can autonomously execute the training curriculum in the real world, the curriculum itself remains task-specific and requires manual design and tuning. Future work should explore more generalizable approaches to real-world curriculum generation. Additionally, our reward design is constrained by the availability of hardware and sensors—for example, RTR lacks access to ground reaction force measurements, which are readily available in simulation but absent in our real-world setup, while instrumenting a force plate beneath the treadmill is a potential workaround. Developing comprehensive real-world sensing methodologies represents a promising direction for advancing continuous robot learning in real-world environments.

 One promising future direction of RTR is to extend to full-scale humanoid robots. Although the current setup cannot handle such robots due to payload constraints, the principle of RTR–a dynamic, force-aware teacher-student system that provides crucial learning curriculum, safety guarantee, and auto resets during real-world learning–is generally applicable. For larger humanoid robots, the robot teacher can be an industrial robot arm or a bridge crane with force sensing, and the latter would also eliminate the need for a treadmill since the crane can move in a large horizontal area.
\section*{Acknowledgment}
The authors would like to express their great gratitude to Yifan Hou for providing valuable help and feedback on the robot arm compliance controller. We thank Sirui Chen, Pei Xu, Kun Lei, Albert Wu, Ruocheng Wang, and Ziang Cao for their input on humanoid reinforcement learning. Finally, we appreciate the helpful discussions from all members of TML and REALab.

This work was supported in part by the NSF Award \#2143601, \#2037101, \#2132519, \#2153854, Sloan Fellowship, and Stanford Institute for Human-Centered AI. We would like to thank Google for the UR5 robot hardware. The views and conclusions contained herein are those of the authors and should not be interpreted as necessarily representing the official policies, either expressed or implied, of the sponsors. 

\clearpage

\bibliography{robots_train_robots}  

\begin{thebibliography}{55}
\providecommand{\natexlab}[1]{#1}
\providecommand{\url}[1]{\texttt{#1}}
\expandafter\ifx\csname urlstyle\endcsname\relax
  \providecommand{\doi}[1]{doi: #1}\else
  \providecommand{\doi}{doi: \begingroup \urlstyle{rm}\Url}\fi

\bibitem[He et~al.(2025{\natexlab{a}})He, Gao, Xiao, Zhang, Wang, Wang, Luo,
  He, Sobanbab, Pan, Yi, Qu, Kitani, Hodgins, Fan, Zhu, Liu, and
  Shi]{he2025asap}
T.~He, J.~Gao, W.~Xiao, Y.~Zhang, Z.~Wang, J.~Wang, Z.~Luo, G.~He, N.~Sobanbab,
  C.~Pan, Z.~Yi, G.~Qu, K.~Kitani, J.~Hodgins, L.~J. Fan, Y.~Zhu, C.~Liu, and
  G.~Shi.
\newblock {{ASAP}}: {{Aligning Simulation}} and {{Real-World Physics}} for
  {{Learning Agile Humanoid Whole-Body Skills}}, Feb. 2025{\natexlab{a}}.

\bibitem[He et~al.(2025{\natexlab{b}})He, Xiao, Lin, Luo, Xu, Jiang, Kautz,
  Liu, Shi, Wang, Fan, and Zhu]{he2025hover}
T.~He, W.~Xiao, T.~Lin, Z.~Luo, Z.~Xu, Z.~Jiang, J.~Kautz, C.~Liu, G.~Shi,
  X.~Wang, L.~Fan, and Y.~Zhu.
\newblock {{HOVER}}: {{Versatile Neural Whole-Body Controller}} for {{Humanoid
  Robots}}, Mar. 2025{\natexlab{b}}.

\bibitem[Radosavovic et~al.(2024{\natexlab{a}})Radosavovic, Xiao, Zhang,
  Darrell, Malik, and Sreenath]{radosavovic2024realworld}
I.~Radosavovic, T.~Xiao, B.~Zhang, T.~Darrell, J.~Malik, and K.~Sreenath.
\newblock Real-world humanoid locomotion with reinforcement learning.
\newblock \emph{Science Robotics}, 9\penalty0 (89):\penalty0 eadi9579, Apr.
  2024{\natexlab{a}}.
\newblock \doi{10.1126/scirobotics.adi9579}.

\bibitem[Radosavovic et~al.(2024{\natexlab{b}})Radosavovic, Kamat, Darrell, and
  Malik]{radosavovic2024learning}
I.~Radosavovic, S.~Kamat, T.~Darrell, and J.~Malik.
\newblock Learning {{Humanoid Locomotion}} over {{Challenging Terrain}}, Oct.
  2024{\natexlab{b}}.

\bibitem[Zhuang et~al.(2024)Zhuang, Yao, and Zhao]{zhuang2024humanoid}
Z.~Zhuang, S.~Yao, and H.~Zhao.
\newblock Humanoid {{Parkour Learning}}.
\newblock In \emph{8th {{Annual Conference}} on {{Robot Learning}}}, Sept.
  2024.

\bibitem[Tobin et~al.(2017)Tobin, Fong, Ray, Schneider, Zaremba, and
  Abbeel]{tobin2017domain}
J.~Tobin, R.~Fong, A.~Ray, J.~Schneider, W.~Zaremba, and P.~Abbeel.
\newblock Domain randomization for transferring deep neural networks from
  simulation to the real world.
\newblock In \emph{2017 {{IEEE}}/{{RSJ International Conference}} on
  {{Intelligent Robots}} and {{Systems}} ({{IROS}})}, pages 23--30, Sept. 2017.
\newblock \doi{10.1109/IROS.2017.8202133}.

\bibitem[Perez et~al.(2017)Perez, Strub, de~Vries, Dumoulin, and
  Courville]{perez2017film}
E.~Perez, F.~Strub, H.~de~Vries, V.~Dumoulin, and A.~Courville.
\newblock Film: Visual reasoning with a general conditioning layer, 2017.
\newblock URL \url{https://arxiv.org/abs/1709.07871}.

\bibitem[Schulman et~al.(2017)Schulman, Wolski, Dhariwal, Radford, and
  Klimov]{schulman2017proximal}
J.~Schulman, F.~Wolski, P.~Dhariwal, A.~Radford, and O.~Klimov.
\newblock Proximal policy optimization algorithms, 2017.
\newblock URL \url{https://arxiv.org/abs/1707.06347}.

\bibitem[Shi et~al.(2025)Shi, Wang, Song, and Liu]{shi2025toddlerbota}
H.~Shi, W.~Wang, S.~Song, and C.~K. Liu.
\newblock {{ToddlerBot}}: {{Open-Source ML-Compatible Humanoid Platform}} for
  {{Loco-Manipulation}}, Feb. 2025.

\bibitem[{Research Nester}(2025)]{mir_mc600_2024}
{Research Nester}.
\newblock Industrial robotic arm market size, share, growth trends, regional
  share, competitive intelligence, forecast report 2025--2037, March 2025.
\newblock URL
  \url{https://www.researchnester.com/reports/industrial-robotic-arm-market/6763}.
\newblock Accessed: 2025-04-28.

\bibitem[Peng et~al.(2018)Peng, Andrychowicz, Zaremba, and
  Abbeel]{peng2018simtoreal}
X.~B. Peng, M.~Andrychowicz, W.~Zaremba, and P.~Abbeel.
\newblock Sim-to-{{Real Transfer}} of {{Robotic Control}} with {{Dynamics
  Randomization}}.
\newblock In \emph{2018 {{IEEE International Conference}} on {{Robotics}} and
  {{Automation}} ({{ICRA}})}, pages 3803--3810, May 2018.
\newblock \doi{10.1109/ICRA.2018.8460528}.

\bibitem[Yuan et~al.(2024)Yuan, Wei, Cheng, Zhang, Chen, and
  Xu]{yuan2024learning}
Z.~Yuan, T.~Wei, S.~Cheng, G.~Zhang, Y.~Chen, and H.~Xu.
\newblock Learning to {{Manipulate Anywhere}}: {{A Visual Generalizable
  Framework For Reinforcement Learning}}, Oct. 2024.

\bibitem[Ha et~al.(2024)Ha, Gao, Fu, Tan, and Song]{ha2024umionlegs}
H.~Ha, Y.~Gao, Z.~Fu, J.~Tan, and S.~Song.
\newblock {{UMI-on-Legs}}: {{Making Manipulation Policies Mobile}} with
  {{Manipulation-Centric Whole-body Controllers}}.
\newblock In \emph{8th {{Annual Conference}} on {{Robot Learning}}}, Sept.
  2024.

\bibitem[Rudin et~al.(2022)Rudin, Hoeller, Reist, and
  Hutter]{rudin2022learning}
N.~Rudin, D.~Hoeller, P.~Reist, and M.~Hutter.
\newblock Learning to {{Walk}} in {{Minutes Using Massively Parallel Deep
  Reinforcement Learning}}.
\newblock In \emph{Proceedings of the 5th {{Conference}} on {{Robot
  Learning}}}, pages 91--100. PMLR, Jan. 2022.

\bibitem[Cheng et~al.(2024)Cheng, Shi, Agarwal, and Pathak]{cheng2024extreme}
X.~Cheng, K.~Shi, A.~Agarwal, and D.~Pathak.
\newblock Extreme {{Parkour}} with {{Legged Robots}}.
\newblock In \emph{2024 {{IEEE International Conference}} on {{Robotics}} and
  {{Automation}} ({{ICRA}})}, pages 11443--11450, May 2024.
\newblock \doi{10.1109/ICRA57147.2024.10610200}.

\bibitem[Tan et~al.(2018)Tan, Zhang, Coumans, Iscen, Bai, Hafner, Bohez, and
  Vanhoucke]{tan2018simtoreala}
J.~Tan, T.~Zhang, E.~Coumans, A.~Iscen, Y.~Bai, D.~Hafner, S.~Bohez, and
  V.~Vanhoucke.
\newblock Sim-to-{{Real}}: {{Learning Agile Locomotion For Quadruped Robots}}.
\newblock In \emph{Robotics: {{Science}} and {{Systems XIV}}}, volume~14, June
  2018.
\newblock ISBN 978-0-9923747-4-7.

\bibitem[Shi et~al.(2023)Shi, Kojio, Makabe, Anzai, Kojima, Okada, and
  Inaba]{shi2023reference}
F.~Shi, Y.~Kojio, T.~Makabe, T.~Anzai, K.~Kojima, K.~Okada, and M.~Inaba.
\newblock Reference-free learning bipedal motor skills via assistive force
  curricula.
\newblock In A.~Billard, T.~Asfour, and O.~Khatib, editors, \emph{Robotics
  Research}, pages 304--320, Cham, 2023. Springer Nature Switzerland.
\newblock ISBN 978-3-031-25555-7.

\bibitem[OpenAI et~al.(2019)OpenAI, Akkaya, Andrychowicz, Chociej, Litwin,
  McGrew, Petron, Paino, Plappert, Powell, Ribas, Schneider, Tezak, Tworek,
  Welinder, Weng, Yuan, Zaremba, and Zhang]{openai2019solving}
OpenAI, I.~Akkaya, M.~Andrychowicz, M.~Chociej, M.~Litwin, B.~McGrew,
  A.~Petron, A.~Paino, M.~Plappert, G.~Powell, R.~Ribas, J.~Schneider,
  N.~Tezak, J.~Tworek, P.~Welinder, L.~Weng, Q.~Yuan, W.~Zaremba, and L.~Zhang.
\newblock Solving {{Rubik}}'s {{Cube}} with a {{Robot Hand}}, Oct. 2019.

\bibitem[Chen et~al.(2023{\natexlab{a}})Chen, Tippur, Wu, Kumar, Adelson, and
  Agrawal]{chen2023visual}
T.~Chen, M.~Tippur, S.~Wu, V.~Kumar, E.~Adelson, and P.~Agrawal.
\newblock Visual dexterity: {{In-hand}} reorientation of novel and complex
  object shapes.
\newblock \emph{Science Robotics}, 8\penalty0 (84):\penalty0 eadc9244, Nov.
  2023{\natexlab{a}}.
\newblock \doi{10.1126/scirobotics.adc9244}.

\bibitem[Chen et~al.(2023{\natexlab{b}})Chen, Wang, {Fei-Fei}, and
  Liu]{chen2023sequential}
Y.~Chen, C.~Wang, L.~{Fei-Fei}, and K.~Liu.
\newblock Sequential {{Dexterity}}: {{Chaining Dexterous Policies}} for
  {{Long-Horizon Manipulation}}.
\newblock In \emph{7th {{Annual Conference}} on {{Robot Learning}}}, Aug.
  2023{\natexlab{b}}.

\bibitem[Chen et~al.(2024)Chen, Wang, Yang, and Liu]{chen2024objectcentric}
Y.~Chen, C.~Wang, Y.~Yang, and K.~Liu.
\newblock Object-{{Centric Dexterous Manipulation}} from {{Human Motion Data}}.
\newblock In \emph{8th {{Annual Conference}} on {{Robot Learning}}}, Sept.
  2024.

\bibitem[Lin et~al.(2025)Lin, Sachdev, Fan, Malik, and Zhu]{lin2025simtoreal}
T.~Lin, K.~Sachdev, L.~Fan, J.~Malik, and Y.~Zhu.
\newblock Sim-to-{{Real Reinforcement Learning}} for {{Vision-Based Dexterous
  Manipulation}} on {{Humanoids}}, Feb. 2025.

\bibitem[Fu et~al.(2024)Fu, Zhao, Wu, Wetzstein, and Finn]{fu2024humanplus}
Z.~Fu, Q.~Zhao, Q.~Wu, G.~Wetzstein, and C.~Finn.
\newblock {{HumanPlus}}: {{Humanoid Shadowing}} and {{Imitation}} from
  {{Humans}}.
\newblock In \emph{8th {{Annual Conference}} on {{Robot Learning}}}, Sept.
  2024.

\bibitem[Gu et~al.(2024)Gu, Wang, Zhu, Shi, Guo, Liu, and
  Chen]{gu2024advancing}
X.~Gu, Y.-J. Wang, X.~Zhu, C.~Shi, Y.~Guo, Y.~Liu, and J.~Chen.
\newblock Advancing {{Humanoid Locomotion}}: {{Mastering Challenging Terrains}}
  with {{Denoising World Model Learning}}.
\newblock In \emph{Robotics: {{Science}} and {{Systems XX}}}. {Robotics:
  Science and Systems Foundation}, July 2024.
\newblock ISBN 9798990284807.
\newblock \doi{10.15607/RSS.2024.XX.058}.

\bibitem[Haarnoja et~al.(2024)Haarnoja, Moran, Lever, Huang, Tirumala, Humplik,
  Wulfmeier, Tunyasuvunakool, Siegel, Hafner, Bloesch, Hartikainen, Byravan,
  Hasenclever, Tassa, Sadeghi, Batchelor, Casarini, Saliceti, Game, Sreendra,
  Patel, Gwira, Huber, Hurley, Nori, Hadsell, and Heess]{haarnoja2024learning}
T.~Haarnoja, B.~Moran, G.~Lever, S.~H. Huang, D.~Tirumala, J.~Humplik,
  M.~Wulfmeier, S.~Tunyasuvunakool, N.~Y. Siegel, R.~Hafner, M.~Bloesch,
  K.~Hartikainen, A.~Byravan, L.~Hasenclever, Y.~Tassa, F.~Sadeghi,
  N.~Batchelor, F.~Casarini, S.~Saliceti, C.~Game, N.~Sreendra, K.~Patel,
  M.~Gwira, A.~Huber, N.~Hurley, F.~Nori, R.~Hadsell, and N.~Heess.
\newblock Learning agile soccer skills for a bipedal robot with deep
  reinforcement learning.
\newblock \emph{Science Robotics}, 9\penalty0 (89):\penalty0 eadi8022, Apr.
  2024.
\newblock \doi{10.1126/scirobotics.adi8022}.

\bibitem[Chebotar et~al.(2019)Chebotar, Handa, Makoviychuk, Macklin, Issac,
  Ratliff, and Fox]{chebotar2019closing}
Y.~Chebotar, A.~Handa, V.~Makoviychuk, M.~Macklin, J.~Issac, N.~Ratliff, and
  D.~Fox.
\newblock Closing the {{Sim-to-Real Loop}}: {{Adapting Simulation
  Randomization}} with {{Real World Experience}}.
\newblock In \emph{2019 {{International Conference}} on {{Robotics}} and
  {{Automation}} ({{ICRA}})}, pages 8973--8979, May 2019.
\newblock \doi{10.1109/ICRA.2019.8793789}.

\bibitem[Ren et~al.(2023)Ren, Dai, Burchfiel, and Majumdar]{ren2023adaptsim}
A.~Z. Ren, H.~Dai, B.~Burchfiel, and A.~Majumdar.
\newblock {{AdaptSim}}: {{Task-Driven Simulation Adaptation}} for {{Sim-to-Real
  Transfer}}.
\newblock In \emph{Proceedings of {{The}} 7th {{Conference}} on {{Robot
  Learning}}}, pages 3434--3452. PMLR, Dec. 2023.

\bibitem[Huang et~al.(2023)Huang, Zhang, Cao, Liu, Xu, Ding, Francis, Chen, and
  Zhao]{huang2023what}
P.~Huang, X.~Zhang, Z.~Cao, S.~Liu, M.~Xu, W.~Ding, J.~Francis, B.~Chen, and
  D.~Zhao.
\newblock What {{Went Wrong}}? {{Closing}} the {{Sim-to-Real Gap}} via
  {{Differentiable Causal Discovery}}.
\newblock In \emph{Proceedings of {{The}} 7th {{Conference}} on {{Robot
  Learning}}}, pages 734--760. PMLR, Dec. 2023.

\bibitem[Schoettler et~al.(2020)Schoettler, Nair, Ojea, Levine, and
  Solowjow]{schoettler2020metareinforcement}
G.~Schoettler, A.~Nair, J.~A. Ojea, S.~Levine, and E.~Solowjow.
\newblock Meta-{{Reinforcement Learning}} for {{Robotic Industrial Insertion
  Tasks}}.
\newblock In \emph{2020 {{IEEE}}/{{RSJ International Conference}} on
  {{Intelligent Robots}} and {{Systems}} ({{IROS}})}, pages 9728--9735, Oct.
  2020.
\newblock \doi{10.1109/IROS45743.2020.9340848}.

\bibitem[Kumar et~al.(2021)Kumar, Fu, Pathak, and Malik]{kumar2021rma}
A.~Kumar, Z.~Fu, D.~Pathak, and J.~Malik.
\newblock {{RMA}}: {{Rapid Motor Adaptation}} for {{Legged Robots}}, July 2021.

\bibitem[Qi et~al.(2022)Qi, Kumar, Calandra, Ma, and Malik]{qi2022inhand}
H.~Qi, A.~Kumar, R.~Calandra, Y.~Ma, and J.~Malik.
\newblock In-{{Hand Object Rotation}} via {{Rapid Motor Adaptation}}.
\newblock In \emph{6th {{Annual Conference}} on {{Robot Learning}}}, Aug. 2022.

\bibitem[Kumar et~al.(2022)Kumar, Li, Zeng, Pathak, Sreenath, and
  Malik]{kumar2022adapting}
A.~Kumar, Z.~Li, J.~Zeng, D.~Pathak, K.~Sreenath, and J.~Malik.
\newblock Adapting {{Rapid Motor Adaptation}} for {{Bipedal Robots}}.
\newblock In \emph{2022 {{IEEE}}/{{RSJ International Conference}} on
  {{Intelligent Robots}} and {{Systems}} ({{IROS}})}, pages 1161--1168, Oct.
  2022.
\newblock \doi{10.1109/IROS47612.2022.9981091}.

\bibitem[Sun et~al.(2021)Sun, Ubellacker, Ma, Zhang, Wang, {Csomay-Shanklin},
  Tomizuka, Sreenath, and Ames]{sun2021online}
Y.~Sun, W.~L. Ubellacker, W.-L. Ma, X.~Zhang, C.~Wang, N.~V. {Csomay-Shanklin},
  M.~Tomizuka, K.~Sreenath, and A.~D. Ames.
\newblock Online {{Learning}} of {{Unknown Dynamics}} for {{Model-Based
  Controllers}} in {{Legged Locomotion}}.
\newblock \emph{IEEE Robotics and Automation Letters}, 6\penalty0 (4):\penalty0
  8442--8449, Oct. 2021.
\newblock ISSN 2377-3766, 2377-3774.
\newblock \doi{10.1109/LRA.2021.3108510}.

\bibitem[Zhang et~al.(2023)Zhang, Ke, Deshpande, Gupta, and
  Srinivasa]{zhang2023cherrypicking}
Y.~Zhang, L.~Ke, A.~Deshpande, A.~Gupta, and S.~Srinivasa.
\newblock Cherry-{{Picking}} with {{Reinforcement Learning}}.
\newblock In \emph{Robotics: {{Science}} and {{Systems XIX}}}, volume~19, July
  2023.
\newblock ISBN 978-0-9923747-9-2.

\bibitem[Lei et~al.(2023)Lei, He, Lu, Hu, Gao, and Xu]{lei2023unio4}
K.~Lei, Z.~He, C.~Lu, K.~Hu, Y.~Gao, and H.~Xu.
\newblock Uni-{{O4}}: {{Unifying Online}} and {{Offline Deep Reinforcement
  Learning}} with {{Multi-Step On-Policy Optimization}}.
\newblock In \emph{The {{Twelfth International Conference}} on {{Learning
  Representations}}}, Oct. 2023.

\bibitem[Xiong et~al.(2024)Xiong, Mendonca, Shaw, and
  Pathak]{xiong2024adaptive}
H.~Xiong, R.~Mendonca, K.~Shaw, and D.~Pathak.
\newblock Adaptive {{Mobile Manipulation}} for {{Articulated Objects In}} the
  {{Open World}}, Jan. 2024.

\bibitem[Jiang et~al.(2024)Jiang, Wang, Zhang, Wu, and
  {Fei-Fei}]{jiang2024transic}
Y.~Jiang, C.~Wang, R.~Zhang, J.~Wu, and L.~{Fei-Fei}.
\newblock {{TRANSIC}}: {{Sim-to-Real Policy Transfer}} by {{Learning}} from
  {{Online Correction}}.
\newblock In \emph{8th {{Annual Conference}} on {{Robot Learning}}}, Sept.
  2024.

\bibitem[Luo et~al.(2025)Luo, Xu, Wu, and Levine]{luo2025precise}
J.~Luo, C.~Xu, J.~Wu, and S.~Levine.
\newblock Precise and {{Dexterous Robotic Manipulation}} via
  {{Human-in-the-Loop Reinforcement Learning}}, Mar. 2025.

\bibitem[Gupta et~al.(2018)Gupta, Mendonca, Liu, Abbeel, and
  Levine]{gupta2018meta}
A.~Gupta, R.~Mendonca, Y.~Liu, P.~Abbeel, and S.~Levine.
\newblock Meta-reinforcement learning of structured exploration strategies,
  2018.
\newblock URL \url{https://arxiv.org/abs/1802.07245}.

\bibitem[Rakelly et~al.(2019)Rakelly, Zhou, Quillen, Finn, and
  Levine]{rakelly2019efficient}
K.~Rakelly, A.~Zhou, D.~Quillen, C.~Finn, and S.~Levine.
\newblock Efficient off-policy meta-reinforcement learning via probabilistic
  context variables, 2019.
\newblock URL \url{https://arxiv.org/abs/1903.08254}.

\bibitem[Yu et~al.(2020)Yu, Tan, Bai, Coumans, and Ha]{yu2020learning}
W.~Yu, J.~Tan, Y.~Bai, E.~Coumans, and S.~Ha.
\newblock Learning fast adaptation with meta strategy optimization, 2020.
\newblock URL \url{https://arxiv.org/abs/1909.12995}.

\bibitem[Peng et~al.(2020)Peng, Coumans, Zhang, Lee, Tan, and
  Levine]{peng2020learning}
X.~B. Peng, E.~Coumans, T.~Zhang, T.-W. Lee, J.~Tan, and S.~Levine.
\newblock Learning agile robotic locomotion skills by imitating animals, 2020.
\newblock URL \url{https://arxiv.org/abs/2004.00784}.

\bibitem[Yu et~al.(2019)Yu, Kumar, Turk, and Liu]{yu2019sim}
W.~Yu, V.~C. Kumar, G.~Turk, and C.~K. Liu.
\newblock Sim-to-real transfer for biped locomotion, 2019.
\newblock URL \url{https://arxiv.org/abs/1903.01390}.

\bibitem[Huang et~al.(2024)Huang, Zhang, Liang, Xu, Kou, Lu, Xu, Xue, and
  Xu]{huang2024mentor}
S.~Huang, Z.~Zhang, T.~Liang, Y.~Xu, Z.~Kou, C.~Lu, G.~Xu, Z.~Xue, and H.~Xu.
\newblock {{MENTOR}}: {{Mixture-of-Experts Network}} with {{Task-Oriented
  Perturbation}} for {{Visual Reinforcement Learning}}, Oct. 2024.

\bibitem[Xu et~al.(2023)Xu, Hu, Doshi, Rovinsky, Kumar, Gupta, and
  Levine]{xu2023dexterous}
K.~Xu, Z.~Hu, R.~Doshi, A.~Rovinsky, V.~Kumar, A.~Gupta, and S.~Levine.
\newblock Dexterous {{Manipulation}} from {{Images}}: {{Autonomous Real-World
  RL}} via {{Substep Guidance}}.
\newblock In \emph{2023 {{IEEE International Conference}} on {{Robotics}} and
  {{Automation}} ({{ICRA}})}, pages 5938--5945, May 2023.
\newblock \doi{10.1109/ICRA48891.2023.10161493}.

\bibitem[Mendonca et~al.(2025)Mendonca, Panov, Bucher, Wang, and
  Pathak]{mendonca2025continuously}
R.~Mendonca, E.~Panov, B.~Bucher, J.~Wang, and D.~Pathak.
\newblock Continuously {{Improving Mobile Manipulation}} with {{Autonomous
  Real-World RL}}.
\newblock In \emph{Proceedings of {{The}} 8th {{Conference}} on {{Robot
  Learning}}}, pages 5204--5219. PMLR, Jan. 2025.

\bibitem[Ha et~al.(2021)Ha, Xu, Tan, Levine, and Tan]{ha2021learning}
S.~Ha, P.~Xu, Z.~Tan, S.~Levine, and J.~Tan.
\newblock Learning to {{Walk}} in the {{Real World}} with {{Minimal Human
  Effort}}.
\newblock In \emph{Proceedings of the 2020 {{Conference}} on {{Robot
  Learning}}}, pages 1110--1120. PMLR, Oct. 2021.

\bibitem[Smith et~al.(2022)Smith, Kostrikov, and Levine]{smith2022walk}
L.~Smith, I.~Kostrikov, and S.~Levine.
\newblock A {{Walk}} in the {{Park}}: {{Learning}} to {{Walk}} in 20 {{Minutes
  With Model-Free Reinforcement Learning}}, Aug. 2022.

\bibitem[Wu et~al.(2023)Wu, Escontrela, Hafner, Abbeel, and
  Goldberg]{wu2023daydreamer}
P.~Wu, A.~Escontrela, D.~Hafner, P.~Abbeel, and K.~Goldberg.
\newblock {{DayDreamer}}: {{World Models}} for {{Physical Robot Learning}}.
\newblock In \emph{Proceedings of {{The}} 6th {{Conference}} on {{Robot
  Learning}}}, pages 2226--2240. PMLR, Mar. 2023.

\bibitem[Smith et~al.(2023)Smith, Cao, and Levine]{smith2023grow}
L.~Smith, Y.~Cao, and S.~Levine.
\newblock Grow your limits: Continuous improvement with real-world rl for
  robotic locomotion, 2023.
\newblock URL \url{https://arxiv.org/abs/2310.17634}.

\bibitem[Bloesch et~al.(2022)Bloesch, Humplik, Patraucean, Hafner, Haarnoja,
  Byravan, Siegel, Tunyasuvunakool, Casarini, Batchelor, Romano, Saliceti,
  Riedmiller, Eslami, and Heess]{bloesch2022towards}
M.~Bloesch, J.~Humplik, V.~Patraucean, R.~Hafner, T.~Haarnoja, A.~Byravan,
  N.~Y. Siegel, S.~Tunyasuvunakool, F.~Casarini, N.~Batchelor, F.~Romano,
  S.~Saliceti, M.~Riedmiller, S.~M.~A. Eslami, and N.~Heess.
\newblock Towards real robot learning in the wild: A case study in bipedal
  locomotion.
\newblock In A.~Faust, D.~Hsu, and G.~Neumann, editors, \emph{Proceedings of
  the 5th Conference on Robot Learning}, volume 164 of \emph{Proceedings of
  Machine Learning Research}, pages 1502--1511. PMLR, 08--11 Nov 2022.
\newblock URL \url{https://proceedings.mlr.press/v164/bloesch22a.html}.

\bibitem[{ROBOTIS}(2024)]{robotis_op3_manual}
{ROBOTIS}.
\newblock {ROBOTIS OP3 e-Manual}.
\newblock \url{https://emanual.robotis.com/docs/en/platform/op3/introduction/},
  2024.
\newblock Accessed: 2025-07-29.

\bibitem[Maples and Becker(1986)]{maples1986experiments}
J.~Maples and J.~Becker.
\newblock Experiments in force control of robotic manipulators.
\newblock In \emph{1986 {{IEEE International Conference}} on {{Robotics}} and
  {{Automation Proceedings}}}, volume~3, pages 695--702, Apr. 1986.
\newblock \doi{10.1109/ROBOT.1986.1087590}.

\bibitem[Hou et~al.(2025)Hou, Liu, Chi, Cousineau, Kuppuswamy, Feng, Burchfiel,
  and Song]{hou2025adaptive}
Y.~Hou, Z.~Liu, C.~Chi, E.~Cousineau, N.~Kuppuswamy, S.~Feng, B.~Burchfiel, and
  S.~Song.
\newblock Adaptive {{Compliance Policy}}: {{Learning Approximate Compliance}}
  for {{Diffusion Guided Control}}, Mar. 2025.

\bibitem[Liang et~al.(2024)Liang, Xu, Hu, Jiang, Huang, and
  Xu]{liang2024makeanagent}
Y.~Liang, T.~Xu, K.~Hu, G.~Jiang, F.~Huang, and H.~Xu.
\newblock Make-{{An-Agent}}: {{A Generalizable Policy Network Generator}} with
  {{Behavior-Prompted Diffusion}}.
\newblock In \emph{The {{Thirty-eighth Annual Conference}} on {{Neural
  Information Processing Systems}}}, Nov. 2024.

\end{thebibliography}
\newpage

\appendix

\section{State and Action Representation}
\label{app:repr}

\subsection{Walking}
Following the RL training setup in~\citet{shi2025toddlerbota}, both the base policy $\pi(\bm{\mathrm{s}}_t)$ and the dynamics-aware policy $\pi(\bm{s}_t, z)$ output $\bm{\mathrm{a}}_t$ as joint position setpoints for proportional-derivative (PD) controllers based on the observable state $\bm{\mathrm{s}}_t$:
\begin{equation}
\bm{\mathrm{s}}_t = \left(\bm{\phi}_t, \bm{\mathrm{c}}_t, \Delta\bm{\mathrm{q}}_t, \bm{\dot{\mathrm{q}}}_t, \bm{\mathrm{a}}_{t-1}, \bm{\omega}_t, \bm{\theta}_t \right),
\label{eq:obs}
\end{equation}
where $\bm{\phi}_t$ is a phase signal, $\bm{c}_t$ represents velocity commands, $\Delta\bm{q}_t$ denotes the position offset relative to the neutral pose $\bm{q}_0$, $\bm{a}_{t-1}$ is the action from the previous time step, $\bm{\omega}_t$ represents the torso’s angular velocity, and $\bm{\theta}_t$ is the torso orientation in euler angles.

In simulation, the critic observation space includes more privileged information to provide better accuracy for the value function, we have:

\begin{equation}
\bm{\mathrm{s}}_t^+ = \left(\bm{\phi}_t, \bm{\mathrm{c}}_t, \Delta\bm{\mathrm{q}}_t, \bm{\dot{\mathrm{q}}}_t, \bm{\mathrm{a}}_{t-1}, \bm{\mathrm{e_q}}_t, \bm{\mathrm{v}}_t, \bm{\omega}_t, \bm{\theta}_t, \bm{\mathrm{m}}_t, \bm{\mathrm{ref}}_t, \tilde{\bm{\mathrm{v}}}_t, \tilde{\bm{\mathrm{\theta}}}_t \right),
\end{equation}

where $\bm{\mathrm{e_q}}_t$ is the error of the current motor position to a reference motor position, $\bm{\mathrm{v}}_t$ is the linear velocity of the robot in the world frame, $\bm{\mathrm{ref}}_t$ is the reference motor pose, while $\tilde{\bm{\mathrm{v}}}_t$ and $\tilde{\bm{\mathrm{\theta}}}_t$ are the linear and angular velocity of the random perturbation applied to the robot. 
In the real world, as some of the privileged observations in the simulation are hard to acquire, we leverage the RTR system's force information, and augment the real-world privileged observation as:

\begin{equation}
\bm{\mathrm{s}}_t^{r+} = \left(\bm{\phi}_t, \bm{\mathrm{c}}_t, \Delta\bm{\mathrm{q}}_t, \bm{\dot{\mathrm{q}}}_t, \bm{\mathrm{a}}_{t-1}, \bm{\mathrm{q}}_t, \bm{\mathrm{v}}_t, \bm{\omega}_t, \bm{\theta}_t, \bm{\mathrm{F}}_t, \bm{\mathrm{\tau}}_t \right),
\end{equation}
where $\bm{\mathrm{q}}_t$ is the joint position of the humanoid, $\bm{\mathrm{v}}_t$ is the linear velocity of the torso measured by a motion tracking system, and $\bm{\mathrm{F}}_t $ and $ \bm{\mathrm{\tau}}_t $ are the force and torque measured by the force-torque sensor mounted on the robot arm teacher.

\subsection{Swing-up}
For the swing-up task, the observation space is defined the same as Equation \ref{eq:obs}. We also include more information to form the task's privileged observation space:

\begin{equation}
\bm{\mathrm{s}}_t^{+} = \left(\bm{\phi}_t, \bm{\mathrm{c}}_t, \Delta\bm{\mathrm{q}}_t, \bm{\dot{\mathrm{q}}}_t, \bm{\mathrm{a}}_{t-1}, \bm{\omega}_t, \bm{\theta}_t, \bm{F}_t, \bm{\tau}_t, \bm{x}_t \right)
\end{equation}

where $\bm{F}_t, \bm{\tau}_t, \bm{x}_t$ are the force reading, torque reading, and arm end effector position, respectively.

\section{FiLM Ablation}
\label{app:film}
In this section, we discuss details regarding the FiLM layers in our sim-to-real adaptation pipeline. As shown in Equation~\eqref{eq:film_layer}, the FiLM layer introduces a scaling and shifting effect to each layer of the hidden network, altering the behavior of the policy network inherently.

One key consideration for the FiLM layer during training is its learning rate, as it decides how fast the FiLM layer will change compared to the policy network. In our implementation, we train the FiLM layers along with the policy network from scratch,  as we find that initializing the policy network from a pretrained model hinders the effect of the FiLM layer over the policy. Recall that:

\begin{equation*}
    \gamma_j^{(i)}, \beta_j^{(i)} = \text{FiLM}_j(z^{(i)}),\quad
    h_j^{(i)} \leftarrow \gamma_j^{(i)} \odot h_j^{(i)} + \beta_j^{(i)}
\end{equation*}
We initialize all the FiLM layers with all-zero weights and all-zero or all-one bias, such that at the start of the training, we have: $\gamma_j^{(i)} = 1.0 , \beta_j^{(i)} = 0.0$, which means the FiLM layers do not affect the network output. As the FiLM layers are trained simultaneously with the policy, they are in a competing condition: the policy network is learning to gain better overall performance given the current FiLM distribution, while the FiLM layers are trained to make use of the environment's latent and maximize the performance in each of the different environments.

We find that a small learning rate for the FiLM layers causes the policy to behave similarly to training without the modulation, limiting its ability to generalize across latent conditions and adapt to different environments. On the other hand, if the learning rate is too large, it will lead to unstable training and lower performance.

\begin{figure}[t]
    \centering
    \includegraphics[width=0.6\linewidth]{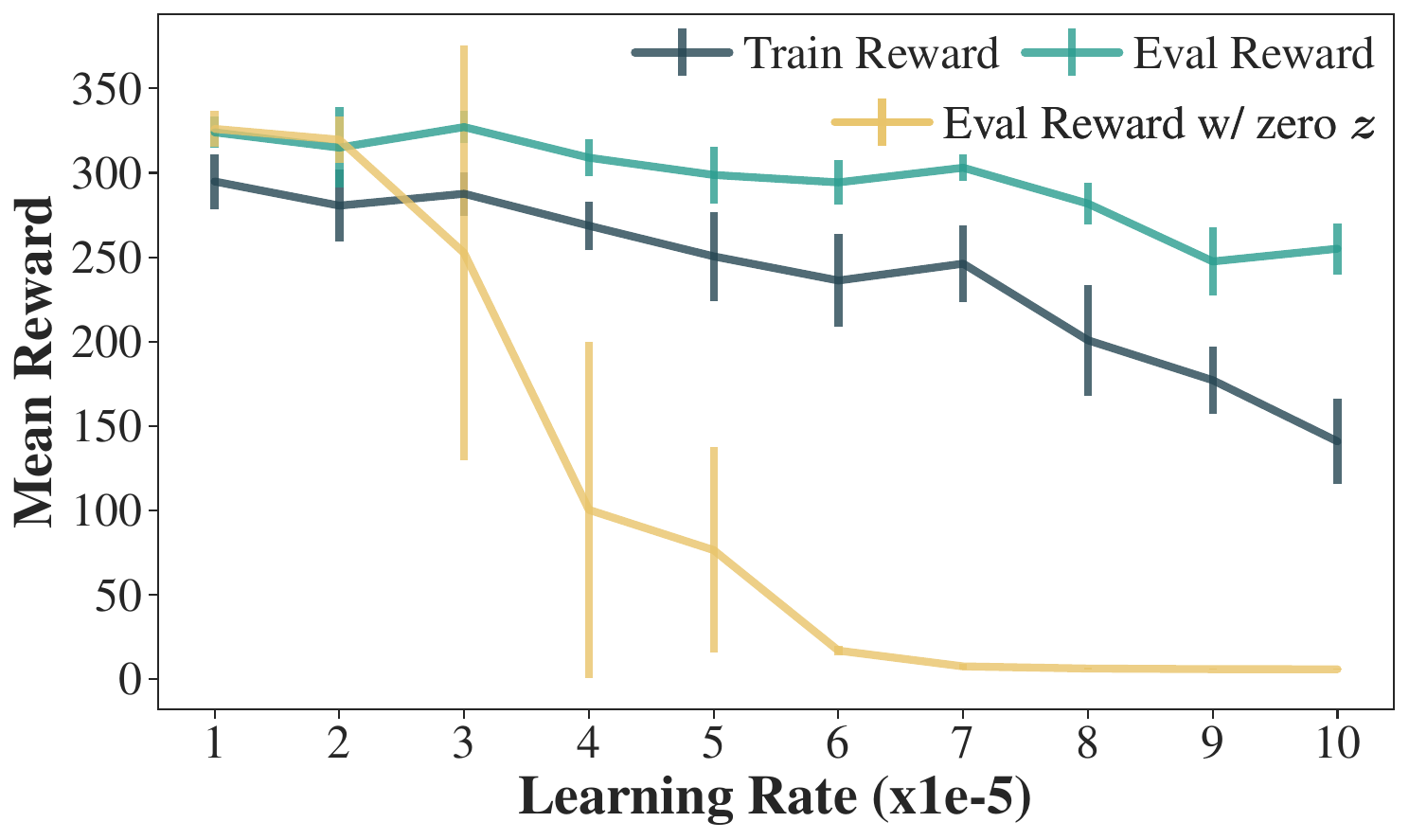}
    \caption{\textbf{FiLM learning rate ablation.} This experiment aims to evaluate the effect of FiLM layer learning rates. All experiments are run under seven seeds. Vertical bars indicate standard deviation.}
    \label{fig:film_lr_ablation}
    \vspace{-2mm}
\end{figure}
We propose a metric to measure the effectiveness for the dynamics latent conditioned $\pi(s, z)$ to utilize the environment information embedded in the latent $z$. Specifically, we compare the performance of the policy given the true latent $z$ computed from the dynamics encoder $z^{(i)} = f_\phi(\mu^{(i)})$ and give an all-zero latent. We find that the performance of an all-zero latent could be even comparable to that given the real z when the FiLM learning rate is small, but it will gradually decrease to near zero when the learning rate increases and the policy starts to learn to use the latent information. On the other hand, the overall performance of the policy will decrease nearly monotonously when the FiLM learning rate increases. We plot the performance curves at different learning rates in Figure~\ref{fig:film_lr_ablation}. For each learning rate, we run seven trials and plot the mean and standard deviation of the performance. We find that the learning rate of $5e-5$ is a good trade-off between performance and the ability to utilize latent information, while $1e-5$ is too small, leading to nearly identical zero latent and true latent performance, and $1e-4$ is too large and leads to poor performance.

\section{Experiment Details}
\subsection{Domain Randomization}
\label{app:dr}
Following the domain randomization settings in~\citep{shi2025toddlerbota}, we slightly increase the domain randomization range to encourage the policy network to better use the dynamics information from the latent. We list the randomized parameters and their range in Table~\ref{tab:dr_range}. We use the encoder-decoder architecture from~\citep{liang2024makeanagent} to encode the physics parameters to a 1024-dimensional latent before sending it to the FiLM layers, though we only use the encoder in our implementation.

\begin{table}[t]
    \centering
    \caption{We randomize the following environment parameters for the walking task.}
    \resizebox{\linewidth}{!}{
    \begin{tabular}{lcccccc}
\toprule
Parameter Names &  Friction &  Damping &  Armature &    Friction Loss  &  Body Mass & EE Mass \\ 
\midrule
Randomize Range & 
  $[0.5, 2.0]$  &  $[0.5, 2.0]$ &  $[0.5, 2.0]$ &   $[0.5, 2.0]$ &  $[-0.3, 0.3]$ & $[0.0, 0.1]$ \\

\bottomrule
    \end{tabular}
    }
    \label{tab:dr_range}
    \vspace{-3mm}
\end{table}

\subsection{Arm Control Policy}
We implement our appliance arm controller based on the open-sourced code of ~\citet{hou2025adaptive}.

For the walking task, we enable appliance control along the XY axis while setting the Z-height directly. We use a stiffness of $[100, 50]$ along the two axis, while setting the damping to $[0.5, 0.5]$ and inertia to $[0.03, 0.03]$, making the arm slightly more compliant on the Y-axis. 

For the swing task, we disable the appliance control for the arm since it makes the phase tracking lag behind. We use position control to let the arm follow the helping or perturbing movement described in Section~\ref{sec:real_world_learning}. During both helping and perturbing modes, each real-world data collection period is evenly divided into 5 bins. In helping mode, 3 bins are randomly selected from the last 4 to apply assistance. In perturbing mode, 1 bin is randomly selected from the first 4 to apply the perturbation.

\subsection{Treadmill Control Policy}
During the walking task, the robot is walking on the treadmill while the treadmill adjusts its speed dynamically to keep the robot in a good position. We use a PD controller for the treadmill speed $v$:

\begin{equation}
    v = v_{base} + k_{p}^1F_x + k_{p}^2\psi
\end{equation}
where $v_{base} = 0.1~\mathrm{m/s}$ is the default treadmill speed, $F_x$ is the force reading along the $x$-axis and $\psi$ is the robot's torso pitch. $k_{p}^1 = 0.2$ and $k_{p}^2 = -5$ are the proportional gain. We set a treadmill speed limit of $0.24~\mathrm{m/s}$ to assure the safety of the humanoid platform.
\subsection{Online Learning Hyperparameters}

During real-world adaptation for walking, we collect a batch of 1024 steps of data before updating the policy on them over 20 epochs using the PPO algorithm with a clipping ratio of 0.2. Both the actor and critic networks are optimized with a learning rate of $1\times10^{-4}$. To encourage exploration, we apply an entropy coefficient of 0.005. 

For training the swing-up task from scratch, we also collect 1024 samples before starting to update the policy. The actor is optimized with a learning rate of $2\times10^{-3}$ and the critic is optimized with a learning rate of $2\times10^{-5}$. To encourage exploration, we apply an larger entropy coefficient of 0.04. Each PPO update is performed over 20 epochs with a clipping ratio of 0.2.

\subsection{Real-world Learning Details}
\section{Comparison with RMA}

\label{app:rma}
Rapad Motor Adaptation (RMA)~\citep{kumar2021rma} is a sim-to-real adaptation method that is similar to our approach. The training of RMA consists of two stages: \textbf{(1)} A pretraining stage, where the policy is trained in a simulated environment with domain randomization. In this stage, the latent information is encoded from the physics parameters by a projection layer, and then concatenated with the observation before being sent to the policy. \textbf{(2)} An adaptation stage, which is to address the problem that the physics parameter of the real world is unknown. During this stage, an adaptation module is trained via supervised learning to reconstruct the latent $z$ from the past observations and actions.

Compared to RMA, our approach bears differences in each stage: for the first stage, our method uses a FiLM layer to modulate the policy network, which is a more flexible and powerful way to utilize the latent information. In contrast, RMA simply concatenates the latent information with the observation. In the second stage, our method does not require the adaptation module to reconstruct the latent $z$, but instead, we optimize a universal latent $z^*$ that is shared across all the environments. This could serve as a good initialization for further training of the optimal real-world latent $z^*_{real}$. On the other hand, RMA tries to reconstruct the latent $z$ from the past observations and actions, which is prone to overfitting, especially when facing the largely out-of-distribution real-world dynamics.

In the following section, we set up simulation and real-world experiments to compare our method with RMA. We try to answer the following questions: \textbf{(1)} How does the FiLM layer modulation affect the performance compared to RMA's simple concatenation? \textbf{(2)} How good is the adaptation module compared to our approach to get a universal latent? For each experiment, we conduct ablation studies and keep all other parameters the same. We then evaluate the fully trained RMA model in the real world to prove that our method is indeed a better choice from each perspective for real-world adaptation.

\subsection{RMA Implementation Details}
We describe the details we use to implement the RMA algorithm on our humanoid platform. 
During phase one training, we randomize the simulation environment using the same parameter range as described in Appendix~\ref{app:dr}, and the randomized parameters are then concatenated to form a physics-information vector. We drop the unchanged digits during the encoding process.

The RMA algorithm is originally designed for a quadruped robot that has relatively low DoFs. Compared with the original implementation, we use a similar process to train a stage one model that takes in the observation and the dynamics latent and outputs the action. During stage two, we make the following changes to adapt the algorithm to our hardware:
(1) We decrease the window length of past observations and actions that used to predict the current dynamics latent from 50 to 15, as the observation of the humanoid platform has higher dimensions, and a too long horizon will cause difficulty in training. This adjustment also aligns with the stack frame length we use during training. (2) We change the 1D convolution layers used to process the past states and actions accordingly, since the context window size is reduced. The new convolution layers now have kernel sizes of (5, 3, 3) and stride steps of (1, 1, 1). Besides these changes, the RMA training reuses the existing real-world learning pipeline, while only optimizing the adaptation module in stage two instead of the dynamics latent as in the implementation of \system.

\subsection{FiLM Latent Modulation}
We first compare the effect of FiLM layer modulation used by \system with the concatenation approach used by the first stage in RMA. To this end, we run three seeds for the RMA to train a dynamics latent conditioned policy in 1024 parallel simulation environments. While the FiLM layer-based policy easily reaches an average evaluation return of higher than 300, the concatenated policy can only reach a return of just over 200. We suspect that this is due to the high dimensionality introduced by concatenation, which will hinder the policy performance. The co-existence of the observation and the dynamics latent in the MLP policy input is also likely to cause the network to ignore the dynamics. The detailed results can be found in the first column of  Table~\ref{tab:rma_ablation}, where the ``Ground Truth'' denotes that both methods have access to the dynamics latent directly predicted from the encoder.

\subsection{Adaptation Module Performance}

During the second stage of RMA, an adaptation module is trained to predict the dynamics latent from past observations. While in \system, a universal latent is trained across randomized environments for initialization of a real-world learning stage. We want to investigate which method could lead to a better estimation of the real dynamics latent. In addition to the design choice of concatenation or FiLM layer modulation studied in the previous subsection, we conduct ablation studies to fully compare the performance of both methods. The results can be found in Table~\ref{tab:rma_ablation}.

In the first row of Table~\ref{tab:rma_ablation}, we use the stage one of RMA to train a dynamics conditioned policy via directly concatenating the dynamics latent to the observation, which will lead to inferior performance than the FiLM layers. We then use both universal latent optimization and the adaptation module to predict the dynamics latent. The result shows that both methods did pretty well to recover the latent, resulting in a nearly identical performance to the true latent used in phase one.

In the second row of Table~\ref{tab:rma_ablation}, we add the dynamics conditioning to the policy via FiLM layers as in \system. The performance of the conditioned policy is higher than that of the concatenated latent. We then compare the two methods' ability to estimate the dynamics latent for the FiLM layers. This time, the RMA-style adaptation module failed to recover a feasible dynamics latent for the FiLM layers, causing the performance to stay at a low position, almost close to that of a random policy. While the prediction error of the adaptation module is decreasing, we suspect this is due to the FiLM layers being more sensitive to the small changes of the dynamics layer, and the adaptation module can't fully close this gap from the past observations and actions.

\begin{table}[t]
    \centering
    \caption{We compare the performance for each stage of \system (ours) and RMA~\citep{kumar2021rma} in this table}
    \resizebox{\linewidth}{!}{
    \begin{tabular}{lccc}
\toprule
 &  Ground Truth &  Adaptation Module (RMA)
&   Universal Latent (RTR)
 \\ 
\midrule
Concatenation (RMA)
 & 
  210.32 $\pm$ 20.75 & 205.74 $\pm$ 15.63 &    207.32 $\pm$ 5.89 \\
FiLM layer (RTR)
 & 
316.10 $\pm$ 11.30 & 10.32 $\pm$ 2.01 & \textbf{305.28 $\pm$ 5.47} \\
\bottomrule
    \end{tabular}
}
\label{tab:rma_ablation}
\vspace{0mm}
\end{table}

\begin{table}[t]
    \centering
    \caption{We compare \system and RMA~\citep{kumar2021rma} in this table, extending the results from Table~\ref{tab:walk_metrics}.}
    \resizebox{\linewidth}{!}{
    \begin{tabular}{lccccc}
\toprule
Method &     Torso Roll $\downarrow$ &     Torso Pitch $\downarrow$ &     EE Force X $\downarrow$ &     EE Force Y $\downarrow$ &     EE Force Z $\downarrow$ \\ 
\midrule
RMA & 
  0.167 $\pm$   0.010 &   0.212 $\pm$   0.006 &   1.172 $\pm$   0.007 &   0.799 $\pm$   0.028 &   2.548 $\pm$   0.216 \\
\system(ours) & 
\textbf{0.093 $\pm$   0.020} & \textbf{0.053 $\pm$   0.044} & \textbf{0.943 $\pm$   0.202} & \textbf{0.754 $\pm$   0.122} & \textbf{0.954 $\pm$   0.445} \\
\bottomrule
    \end{tabular}
    }
    \label{tab:rma_metrics}
    \vspace{-3mm}
\end{table}

\subsection{Real-World Performance}
Finally, we run the RMA model in the real world and test its performance using the same metrics as the main paper. While our implementation of the RMA method successfully adapts to the real world to produce a walking behavior somewhat similar to the simulation, the walking gait of the RMA policy is not that stable, and the humanoid often leans forward and is frequently recovered by the \system hardware. The metrics in Table~\ref{tab:rma_metrics} show that our method clearly outperforms RMA for real-world adaptation.

In summary, during the first stage of training, the approach of using the FiLM layer in \system outperforms the concatenation method used by RMA. During the second stage, our approach of universal latent optimization could provide similar initial latent compared to RMA, and is more stable when combined with the FiLM layers. What's more, our approach of universal latent optimization leads to a third real-world tuning stage, and has the ability to further boost the real-world performance with the \system hardware. The results in both simulation and real-world experiments lead to the conclusion that the dynamics latent optimization pipeline of \system is better suited for sim-to-real adaptation of humanoids than the approach used by RMA.

\end{document}